# What can computational models learn from human selective attention? A review from an audiovisual crossmodal perspective

**Di Fu** [1,2,3], **Cornelius Weber** [3], **Guochun Yang** [1,2], **Matthias Kerzel** [3], **Weizhi Nan** [4], **Pablo Barros** [3], **Haiyan Wu** [1,2], **Xun Liu** [1,2,*], **and Stefan Wermter** [3]

[1] *CAS Key Laboratory of Behavioral Science, Institute of Psychology, Beijing, China*
[2] *Department of Psychology, University of Chinese Academy of Sciences, Beijing, China*
[3] *Department of Informatics, University of Hamburg, Hamburg, Germany*
[4] *Department of Psychology and Center for Brain and Cognitive Sciences, School of Education, Guangzhou University, Guangzhou, China*

Correspondence*:
Xun Liu
liux@psych.ac.cn

## ABSTRACT

Selective attention plays an essential role in information acquisition and utilization from the environment. In the past 50 years, research on selective attention has been a central topic in cognitive science. Compared with unimodal studies, crossmodal studies are more complex but necessary to solve real-world challenges in both human experiments and computational modeling. Although an increasing number of findings on crossmodal selective attention have shed light on humans' behavioral patterns and neural underpinnings, a much better understanding is still necessary to yield the same benefit for computational intelligent agents. This article reviews studies of selective attention in unimodal visual and auditory and crossmodal audiovisual setups from the multidisciplinary perspectives of psychology and cognitive neuroscience, and evaluates different ways to simulate analogous mechanisms in computational models and robotics. We discuss the gaps between these fields in this interdisciplinary review and provide insights about how to use psychological findings and theories in artificial intelligence from different perspectives.

## 1 INTRODUCTION

*"The art of being wise is knowing what to overlook." – William James, 1842-1910.*

The real world is complex, uncertain and rich of dynamic ambiguous stimuli. Detecting sudden changes in the environment is significant for organisms to survive because these events need prompt identification and response (Todd and Van Gelder, 1979). Considering the limited capacity for processing information, selective attention is like a filter with the ability to remove unwanted or irrelevant information and thus optimizes a human's action to achieve the current goal (Desimone and Duncan, 1995). It is crucial as well for intelligent agents to integrate and utilize external and internal information efficiently and to reach a signal-to-noise ratio as high as humans can (signal detection theory, SDT) (Green et al., 1966; Swets, 2014).

Selective attention is involved in the majority of mental activities, it is used to control our awareness of internal mind and of the outside world and it helps us integrate the information from multidimensional and multimodal inputs. Eventually, those cognitive processes also contribute to different aspects of consciousness (Posner and Rothbart, 1998). Selective attention is predominantly categorized by psychologists and neuroscientists into "endogenous" and "exogenous" attention. Endogenous attention helps to allocate limited cognitive resources to the current task (Corbetta and Shulman, 2002; Posner and Snyder, 1975; Styles, 2006). The metaphor for this process is described as directing a spotlight in a dark room. Such a process helps us for instance to search for one specific email only by glimpsing the crammed email box. However, the action can sometimes be interrupted by attractive advertisements or breaking news on a website. This latter kind of orienting attention is called exogenous attention which is usually caused by an unexpected change in the environment. It is considered to be instinctive and spontaneous and often results in a reflexive saccade (Smith et al., 2004; Styles, 2006). Another point of view distinguishes between "covert" and "overt" orienting attention: covert attention can attend events or objects with the absence of eyes movement, while overt attention guides the fovea to the stimulus directly with eyes or head movements (Posner, 1980). This is because covert attention requires inhibition of saccades to sustain fixation, which is not needed during overt attention (Kulke et al., 2016).

To understand the mechanisms underlying selective attention is helpful for computational models of selective attention for different purposes and requirements (Das et al., 2017). However, theories in the field of human selective attention are complex and non-unified. Some theories are metaphysical and mystifying, especially for readers that lack experience in humans' behavioral and neural studies. Frintrop et al. (2010) published a survey about computational visual systems with an extensive description of the concepts, theories and neural pathways of visual attention mechanisms. It is stated that "the interdisciplinarity of the topic holds not only benefits but also difficulties: concepts of other fields are usually hard to access due to differences in vocabulary and lack of knowledge of the relevant literature." These interdisciplinary challenges are still unsolved thus far. Additionally, the development and application of technical measurements and methods like functional magnetic resonance imaging (fMRI), Magnetoencephalography (MEG), and state-of-the-art artificial neural networks (ANN) and deep learning (DL) open up a new window for studies on humans, primates, and robots. Considerable new findings should be summarized and added into the current framework.

Although there are several review articles on selective attention in the field of both psychology and computer science (Frintrop et al., 2010; Shinn-Cunningham, 2008; Lee and Choo, 2013), most of them only focus either on a single modality or on general crossmodal processing (Lahat et al., 2015; Ramachandram and Taylor, 2017). However, it is essential to combine and compare selective attention mechanisms from different modalities together to provide an integrated framework with similarities and differences among various modalities. In the current review, firstly, we aim to integrate selective attention concepts, theories, behavioral, and neural mechanisms studied by the unimodal and crossmodal experiment designs. Secondly, we aim to deepen the understanding of the interdisciplinary work in multisensory integration and crossmodal learning mechanisms in psychology and computer science. Thirdly, we aim to bridge the gap between humans' behavioral and neural patterns and intelligent system simulation to provide theoretical and practical benefits to both fields.

The current review is organized into the following parts. Section 2 is about the existing mainstream attention theories and models based on human experimental findings and attention mechanisms in computer science. Section 3 summarizes human visual selective attention studies and introduces the modeling work in computer science inspired by psychology. Section 4 describes results on less studied auditory selective attention and the corresponding modeling work. Section 5 reviews mechanisms and models about crossmodal selective attention and state-of-the-art approaches in intelligent systems. Here, to provide focus, we select the most representative phenomena and effects in psychology: Pop-out Effect (visual attention), Cocktail Party Effect (auditory attention), and audiovisual crossmodal integration and conflict resolution (crossmodal attention). Since these effects are also well-established and often simulated in computer science, we highlight the classic and latest work. Finally, we discuss the current limitations and the future trends of utilization and implication of humans' selective attention in artificial intelligence.

## 2 DIFFERENT THEORIES AND MODELS OF SELECTIVE ATTENTION

### 2.1 Classic bottom-up and top-down control *vs.* Priority map theory

The mainstream view of selective attention proposes that there exist two complementary pathways in the brain cortex, the dorsal and ventral systems. The former, which includes parts of the intraparietal sulcus (IPS) and frontal eye field (FEF), is in charge of the top-down process guided by goals or expectations. The latter, which involves the ventral frontal cortex (VFC) and right temporoparietal junction (TPJ), is in charge of the bottom-up process triggered by sensory inputs or salient and unexpected stimuli without any high-level feedback. When novelty is perceived, the connection between the TPJ and IPS plays the role of cutting off continuous top-down control (Corbetta and Shulman, 2002) (see **Figure 1a**). The classic bottom-up and top-down control theory can explain many cases in selective attention, and lots of computational models are based on this simple theoretical structure (Mahdi et al., 2019; Fang et al., 2011). However, in some cases, for instance, stimuli with the same physical salience but associated with reward (Anderson et al., 2011) or emotional information (Vuilleumier, 2005; Pessoa and Adolphs, 2010) can also capture attention, even if not correlated with the current goal. Thus, beyond the classical theoretical dichotomy, the priority map theory remedies the explanatory gap between goal-driven attentional control and stimulus-driven selection by adding the past selection history to explain some strong selection biases (Awh et al., 2012). For example, a novice mother who has recently delivered a child is more easily and frequently attracted by pregnant women and babies on the street, even though the "pregnancy history" is irrespective of current goals and physical salience. In general, these two theoretical frameworks are helpful to explain most behavioral cases of selective attention.

### 2.2 Functional neural networks model

Fan and Posner distinguished three interrelated attention neural networks with different functional components in the human brain: alerting, orienting, and executive control (Fan et al., 2005, 2002; Fan and Posner, 2004). They designed the Attentional Network Test (ANT) by combining the classic Flanker task and Posner cueing task to provide a quantitative measurement for studying each component. The component of the alerting network increases the focus on the potential stimuli of interest, and anatomical mechanisms of alerting are correlated with the thalamic, frontal, and parietal regions. The orienting function is responsible for selecting task-related or survival-related information from all the sensory inputs. The orienting network also determines an attention shift between exogenous attention engagement (bottom-up) and endogenous attention disengagement (top-down). Orienting is associated with the superior parietal lobe (SPL), TPJ and frontal eye fields (FEF). The executive control component of attention plays a dominant role in planning, decision-making, conflict detection and resolution. The anterior cingulate cortex (ACC) and lateral prefrontal cortical regions are involved in the executive control component (Benes, 2000). The main contribution of functional neural networks model and ANT is separating attention into clear sub-component. Clinical studies using the ANT can explore the specific difference of cognitive performance between patients and healthy participants (Togo et al., 2015; Urbanek et al., 2010). For example, Johnson et al. (2008) used the ANT to test children with attention deficit hyperactivity disorder (ADHD) and found that they show deficits in the alerting and executive control networks but not in the orienting network. Their results can provide evidence and direction for clinical treatment.

## 2.3 Neural oscillation model

Neural oscillations characterize the electrical activity of a population of neurons (Musall et al., 2012). Synchronization of oscillations is the coordination of firing patterns of groups of neurons from different brain areas (Varela et al., 2001). In contrast, desynchronization of oscillations is the inhibition of neuron activities with opposite phases. Synchronization and desynchronization can occur simultaneously and contribute to various cognitive processes. Clayton and colleagues (2015) propose a gamma-theta power-phase coupling model of attention and point out that attention is selectively adjusted via the excitation of task-relevant processes and the inhibition of task-irrelevant processes (see **Figure 1b**). The excitation of task-relevant processes is controlled by frontomedial theta (fm-theta) power (4-8 Hz) from the posterior medial frontal cortex (pMFC) to the lateral prefrontal cortex (LPFC). Among the communication between LPFC and excited sensory areas, gamma power (>30 Hz) is associated with the excitation of the task-relevant processes. The inhibition of task-irrelevant processes is linked with alpha power (8-14 Hz). The pMFC deploys the crucial inhibition processing by controlling the default mode network (posterior cingulate cortex (PCC) and ventromedial prefrontal cortex (vmPFC)) via the alpha oscillation. The limitation of the model is that the results obtained and presented across different brain regions are mainly correlations and descriptive results rather than causal relationships. Besides, most of the empirical evidence for the model was obtained by visual tasks instead of other modalities. Nevertheless, this gamma-theta power-phase coupling model shows interpretative neural pathways of the neural oscillation of selective attention.

## 2.4 Free-energy model and information theory

The free-energy model inspired by the Helmholtz machine (Dayan et al., 1995) and Bayesian surprise (Itti and Baldi, 2009) explains attention from a hierarchical inference perspective (Feldman and Friston, 2010; Friston, 2009). The gist of the model is that the stimuli in the living environment can be viewed as sensory inputs, surprise or uncertainty which can increase the entropy of the human brain. However, our brains have a tendency to maintain the information order to minimize the energy cost caused by surprise. Humans cannot easily switch off the input signal channels by following their own inclinations. In doing so, perception brings about the sensory inputs, and attention infers the consequence caused by the inputs to adjust action and control the entropy growth. In other words, attention is the synaptic gain encoding the precision of inferring prediction errors during this Bayes-optimal scheme.

Corresponding to the free-energy model, Fan's review (2014) tries to combine the information theory and experimental neural findings to explain the top-down mechanisms of humans' cognition control (the hub of the cognition capacity) and selective attention. Inspired by the free-energy view, Fan points out that cognitive control is a high-level uncertainty or entropy reduction mechanism instead of a low-level automatic information perception. According to Shannon's information theory (Shannon, 1948), uncertainty can be quantified by entropy, and the rate of entropy is used to calculate the time density of the information transmission through different channels. Performance costs appear during cognitive channel switching. The benefits of the information theory are that attention or other cognitive processes can be quantified, and situations (like incongruent or congruent conditions in conflict processing) can be computed as bits quantitatively. Fan assimilates stimulus types, time frequency of the stimulus presentation, and human reaction time from cognitive psychology experimental tasks into entropy, surprise, and channel capacity. In this theory, if we know the probability of an event or a stimulus condition, we can calculate the surprise value of that condition and infer the information processing rate. For example, studies found that visual attention can select 30-60 bits per glimpse (Verghese and Pelli, 1992) and the upper limit of human information processing is around 50 bps. Under this framework, the anterior insula (AI) and the anterior cingulate cortex (ACC) are associated with processing the uncertain inputs and the frontoparietal cortex plays a ubiquitous role in the active control.

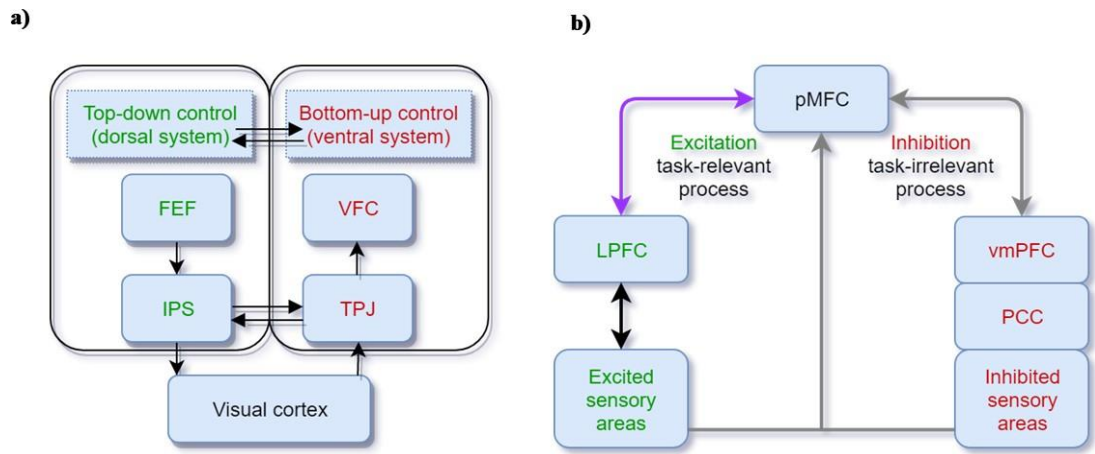

**Figure 1. a)** Neuroanatomical model of bottom-up and top-down attentional processing in the visual cortex. The dorsal system (green) executes the top-down attentional control. FEF: frontal eye field; IPS: intraparietal sulcus. The ventral system (red) executes the bottom-up processing. VFC: ventral frontal cortex; TPJ: temporoparietal junction (adapted from Corbetta and Shulman (2002)); **b)** Cortical oscillation model of attentional control in visual and auditory sensory areas. The posterior medial frontal cortex (pMFC) modulates the selective attention by the excitation of task-relevant processing and the inhibition of task-irrelevant processing. Theta oscillations facilitate the communication between the pMFC and lateral prefrontal cortex (LPFC) (purple arrow). Gamma oscillations and alpha oscillations are promoted in task-relevant and task-irrelevant cortical areas respectively (grey arrows) (adapted from Clayton et al. (2015)).

Entropy increase and decrease happen in our daily life. For instance, undone items on our to-do lists make us feel stressful discomfort, like a full hard disk with compressed working space. There is an impulsion to wipe out tasks to get the peace and control of our mind. Research from network neuroscience takes a similar viewpoint that the brain is designed to be functioning with the lowest cost (Barbey, 2018; Bullmore and Sporns, 2012). However, the free-energy model and information theory concentrate on top-down control pathways which may fail to explain some bottom-up phenomena. For instance, why can human attention be captured by the external salient stimuli involuntarily? It can cause the rise of the information entropy and be opposite to the hypothesis that the human brain instinctively resists the disorder. Another unclear question is to which degree processing channels are good models of neural information processing pathways.

## 2.5 Attention mechanisms in computer science

Attention models have been proposed and applied in computer science for decades, and the concept of attention mechanisms has been growing in recent years in terms of its high performance on sequence modeling research. Bio-inspired implementations of attention in computer science address the limited computation capacity of machines through assigning computational resources by priority. Previous models (1980s - 2014) mainly use the saliency-based winner-take-all algorithm based on human datasets to mimic humanlike visual or auditory attention (Borji and Itti, 2012; Lee and Choo, 2013). Those models aim to extract the target information from the environment or noisy background. In recent years since 2014, attention mechanisms have been applied to Convolutional Neural Networks (CNNs), Recurrent Neural Networks (RNNs), and Long-short Term Memory (LTSM) for sequence modeling work. Attention mechanisms were firstly used in computer vision (Ba et al., 2014) and then became widely used across different domains according to the type of input data, such as object recognition (Hara et al., 2017), image description generation (Xu et al., 2015), speech recognition (Chorowski et al., 2015), machine translation (Luong et al., 2015), video caption generation (Gao et al., 2017b), sentiment classification (Wang et al., 2016), visual question answering (Li et al., 2018), etc.

Attention mechanisms in computer science can be distinguished as soft and hard attention (Xu et al., 2015), or as global and local attention (Luong et al., 2015). Soft attention is the expectation of selected information in the input attention distribution. For example, in machine translation, attention scores mean different weights assigned to words in the source sentence according to each word in the target sentence. Attention can be further separated into item-wise and location-wise soft attention. Soft attention focuses more broadly than hard attention. Hard attention only concentrates on information of the specific location by assigning zero weight to other information (Xu et al., 2015). The concepts of global and local attention vaguely correspond to soft and hard attention respectively. Moreover, an important application is the self-attention mechanism (Vaswani et al., 2017). It can be applied in each decoder layer of neural networks to achieve distributed processing. Besides, self-attention can capture features of long-range dependencies both in the source and target text to learn the inherent structure of inputs (Bahdanau et al., 2014). In this way, self-attention shows good performance when the input sentence is too long as in machine translation (Luong et al., 2015) or the input image is too large as in computer vision (Peng et al., 2019). Therefore, the self-attention mechanism can accelerate training speeds and improve performances of sequence modeling work.

In summary, we conclude in this section that human attention is a process to allocate cognitive resources with different weights according to the priority of the events. Similarly, in computer science, attention mechanisms in models are designed to be allocating different weights to relevant input information and ignore irrelevant information with low-valued weights. However, the connection between computer science models and psychology is still loose and broad. Especially for understanding crossmodal selective attention from a functional view, it is required to explore the human cognition processing from a computational perspective, which is also beneficial for confirming psychological and biological hypotheses in computer science.

## 3 VISUAL SELECTIVE ATTENTION - "POP-OUT" EFFECT

### 3.1 Behavioral and neural mechanisms

Many systematic reviews in the areas of primate vision and computer vision have introduced the concepts and research findings in visual selective attention (Borji and Itti, 2012; Frintrop et al., 2010; Lee and Choo, 2013). In our current review, we further concentrate in particular on mechanisms of the "pop-out" effect and computational models based on the saliency map. The "pop-out" effect usually happens when an object has more salient physical features than other objects in the context, such as location, color, shape, orientation, brightness, etc. (VanRullen, 2003). Saliency can also be extended to affective and social domains, like familiarity, threat, etc. (Fan, 2014). Humans' attention can be immediately captured by salient objects, which can explain why the warning signs on streets are always red and apparent. In general, the "pop-out" processing equals to saliency processing. Itti and Baldi (2009) point out that a salient object is not salient by itself but contains increased uncertainty or more entropy compared with the environment. The difference between prior belief of the real world and posterior observation by an observer (Bayesian surprise) captures human attention.

Nevertheless, controversy remains about the role of top-down control when a salient stimulus captures attention. Stimulus-driven theory (bottom-up saliency hypothesis) suggests that an abrupt-onset object can automatically capture humans' attention without any intention and be processed faster than other non-onset elements (Theeuwes, 1991; Yantis and Jonides, 1984). To the contrary, the goal-driven theory (Bacon and Egeth, 1994) and the contingent capture hypothesis (Folk et al., 1992) propose that the overlap dimension between stimulus properties and task setting goals is the crucial factor, since it can determine whether the salient stimulus can be captured or not. Experiments show that if the salient stimulus has no relevant feature, participants adopt a feature-search mode autonomously to suppress the distraction from the salient stimulus (Bacon and Egeth, 1994).

Hybrid theories attempt to integrate components of both stimulus-driven and goal-driven theories in attention capture. Findings from monkey studies showed that attention selection through biased competition occurred when the target and the distractor were both within the receptive field. Neurons responded primarily to the target, whereas the responses to the distractor were attenuated (Desimone and Duncan, 1995). Subsequently, Mounts (2000) discovered a phenomenon named "surround inhibition". If a salient stimulus appears near the target, it can be inhibited by a top-down control. Later, the signal suppression hypothesis proposed that the salient stimulus automatically generates a salience signal at first and then the signal can be subsequently suppressed, possibly resulting in no attention capture (Gaspelin et al., 2015, 2017; Sawaki and Luck, 2010) (the theories are summarized in **Table 1**).

**Table 1.** Main theories of visual selective attention based on various processing pathways

| Theory | Viewpoint | Processing |
|---|---|---|
| Stimulus-driven Theory (1992) | Singletons automatically capture visual attention | Bottom-up |
| Goal-driven Theory (1992) | Individuals' intentions determine attentional capture | Top-down |
| Contingent Capture Hypothesis (1992) | Contingent on attentional control settings induced by task demands | Top-down |
| Attention Selection Bias Competition (1995) | Response to distractors around the target is inhibited | Bottom-up & Top-down |
| Signal Suppression Hypothesis (2010) | Salience signal automatically generated by singletons can be suppressed | Bottom-up & Top-down |

Neural findings of humans and primates contribute a lot to understand saliency processing in the primary cortex and subcortex. The saliency map theory (Li, 1999, 2002) suggests that neurons in the primary visual cortex (V1) play a crucial role for the input feature processing during the "pop-out" effect. V1 is the neural foundation of the preattentive process during visual search, and it only responds to stimuli located in the classical receptive fields (CRFs). In this saliency map theory, V1 is considered to define the saliency degree of visual inputs. Various features of the target and context enter into the V1 CRFs at the same time. When features of the target are more significant than the context, the target pops out. The saliency map computes the saliency value for all locations in the CRFs rather than only encoding the target location (Veale et al., 2017). In comparison to the classical feature integration model (Treisman and Gormican, 1988) and Itti's saliency model (Itti and Koch, 2000), the main property of the saliency map theory is that saliency processing is only based on a single general feature selection map rather than using a combination map to bind several individual feature maps together. Furthermore, dominant inputs from V1 convey signals to an evolutionarily old structure in the midbrain – the super colliculus (SC). Superficial layers of the SC encode saliency representations through centre-surround inhibition and transfer the inputs to deep layers to trigger priority selection mechanisms to guide attention and gaze (Veale et al., 2017; White et al., 2017; Stein et al., 2002). There is not only bottom-up processing in the primary visual cortex and SC, but also top-down processing. Within the primary visual cortex, the top-down mechanism is mediated by V2 and the interaction occurs in human V4 (Melloni et al., 2012). Moreover, deep layers of the SC represent goal-related behaviors independent of the visual stimuli (Hafed et al., 2008; Hafed and Krauzlis, 2008; Veale et al., 2017).

The large-scale human brain networks also play important roles in visual selective attention. The salience network (SN), composed of AI (anterior insula) and ACC (anterior cingulate cortex), is considered to be working as the salience filter to accept inputs from the sensory cortex and trigger cognitive control signals to the default mode network (DMN) and central-executive network (CEN). Functions of the SN are mainly about accomplishing the dynamic switch between externally and internally oriented attention (Menon and Uddin, 2010; Uddin, 2015; Uddin and Menon, 2009). Another taxonomic cingulo-opercular network shares a large overlap with the SN, containing the anterior insular/operculum, dorsal anterior cingulate cortex (dACC), and thalamus. The cingulo-opercular network has the highest cortical nicotinic acetylcholine receptor (nACHr) density, which is highly correlated with attention functions (Picard et al., 2013). This network shows more elaborate functions in selective attention (Dunlop et al., 2017). However, conclusions about functions of the cingulo-opercular network are not consistent. For instance, Sadaghiani and colleagues (2014) revealed that the cingulo-opercular network plays a role in staying alert but not in selective attention during visual processing. In sum, the V1 and SC consist of primary cortex-subcortex pathways of saliency processing and attention orienting. The AI and ACC consist of large-scale functional networks of saliency processing, alertness and attention shifting. However, the correlation or interaction between these two pathways remains unclear.

Besides elementary physical salient features, the meaning map generated by human eye movements viewing real-world scenes also proved to play a critical role in attentional guidance. Henderson and colleagues (2017) encode the meaning maps comparable to the image salience and operationalize the attention distribution to be duration-weighted fixation density. Their work demonstrates that both, salience and meaning, predict attention but only meaning guides attention while viewing real-world scenes. According to the cognitive-relevance theory of attentional guidance, the meaning maps contain more semantic information for the real context. Their updated findings appear to be particularly insightful and practical for artificial intelligence methods for labeling real-world images.

## 3.2 Computational models

Based on human saccade and fixation research, a vast body of bio-inspired visual attention models has been developed and broadly applied in object segmentation (Gao et al., 2017a), object recognition (Klein and Frintrop, 2011), image caption generation (Bai and An, 2018), and visual question answering (VQA) (Liu and Milanova, 2018). Consistent with humans' visual processing pathways, models in visual attention are generally classified based on the bottom-up and top-down streams (Borji and Itti, 2012; Liu and Milanova, 2018). Bottom-up models are successful in modeling low-level and early processing stages (Khaligh-Razavi et al., 2017). The most classic saliency model, which uses features of color, orientation, edge, and intensity, allocates an attention weight to each pixel of the image (Itti and Koch, 2000; Itti et al., 1998) (see **Figure 4a**). Borji and Itti (2012) point out that the attention model aims to predict the human eye fixation with minimal errors. The "winner-take-all" strategy is the core algorithm of saliency models. However, several criticisms on the saliency model cannot be ignored either. A salient location obtained from the center-surround scheme only simply corresponds to a pixel of an image scene with higher contrast but not the whole or a part of an object (Lee and Choo, 2013; VanRullen, 2003) (also see **Figure 4a**).

In contrast, high-level task-driven attention models remain to be explored and developed further. Some research predicts human eye fixation with free-viewing scenes based on end-to-end deep learning architectures (Kruthiventi et al., 2017; Jetley et al., 2016; Kummerer et al., 2017). Deep neural networks (DNNs) have sometimes been shown to have better performance than other known models by using top-down processing mechanisms. Especially, DNNs can successfully simulate human-like attention mechanisms (Hanson et al., 2018). Task-driven components here can be implemented as prior knowledge, motivation, and other types of cues except for targets. Furthermore, models like DeepFeat incorporating bottom-up and top-down saliency maps by combining low- and high-level visual factors surpass other individual bottom-up and top-down approaches (Mahdi et al., 2019). Nowadays, computer vision research intends to make models learn the semantic meaning rather than simply classify objects. For instance, image captioning requires models not only to detect objects but also extract relationships between objects (Hinz et al., 2019). Co-saliency tends to be a promising preprocessing step for many high-level visual tasks such as video foreground extraction, image retrieval, and object detection. Because co-saliency implies priorities based on human visual attention, it can detect the most important information among a set of images with a reduced computational demand (Yao et al., 2017). In future research, co-saliency approaches may be combined with the meaning maps of human attention for better image interpretation accuracy.

As the number of interdisciplinary studies keeps increasing, research from psychology and artificial intelligence complement each other deepening the understanding of human visual attention mechanisms and improving the performance of computational models. On the one hand, psychologists interpret humans' behavioral or neural patterns by comparing them with performance of DNNs. For example, Eckstein and colleagues (2017) found that human participants often miss giant targets in scenes during visual search but computational models such as Faster R-CNN (Ren et al., 2015), R-FCN (Dai et al., 2016), and YOLO (Redmon and Farhadi, 2017) do not show any similar recognizing failures. Their results suggest that humans use "missing giant targets" as the response strategy to suppress potential distractors immediately. On the other hand, computer scientists interpret features of computational models by comparing their performance in simulations of humans' behaviors. For instance, Hanson and colleagues (2018) found that the Deep Learning (DL) network rather than the single hidden layer backpropagation neural network can replicate human category learning. This is because hidden layers of the DL network can selectively attend to relevant category features as humans do during category learning.

## 4 AUDITORY SELECTIVE ATTENTION – COCKTAIL PARTY EFFECT

### 4.1 Behavioral and neural mechanisms

At a noisy party, a person can concentrate on the target conversation (a top-down process) and easily respond to someone calling his/her name (a bottom-up process). This capability (in a real-life scenario) is named "Cocktail Party Effect" (Cherry, 1953). Auditory information conveys both temporal and spatial features of objects. For instance, we can determine whether water in a kettle is boiling by the special sounds of different heating phases. Auditory scene analysis (ASA) allows the auditory system to perceive and organize sound information from the environment (Bregman, 1994). Since humans cannot close their ears spontaneously to avoid irrelevant information, selective attention is important to segregate forefront auditory information from a complex background and distinguish meaningful information from noise. Besides, auditory selective attention allows humans to localize sound sources and filter out irrelevant sound information effectively.

In the Cocktail Party problem, energetic masking and informational masking cause the ambiguity between the auditory target and noise in the environment. Energetic masking occurs when different sound sources have overlaps in frequency spectra at the same time. The perception and recognition of the target sound can be weakened physically by noise. Informational masking occurs when the target and masker voices sound similar. The listener cannot discriminate them perceptually (Brungart, 2001; Lidestam et al., 2014). The neural mechanisms of these two causes are different. Scott et al. (2004) asked participants to listen to a target speaker with added noise (energetic masking) or added speech (informational masking). They found that informational masking was associated with the activation in the bilateral superior temporal gyri (STG) and energetic masking was associated with the activation in the frontoparietal cortex. The activation was correlated with explicit attentional mechanisms but not specifically to the auditory processing.

In accordance with the Gestalt framework, ASA is the solution to the Cocktail Party problem (Bee and Micheyl, 2008). Similar to visual processing, ASA can be separated into two components. The primitive analysis (bottom-up process) and the schema-based processing (top-down process) (Bregman, 1994). In the primitive analysis, auditory signals are separated into independent units and integrated into disparate auditory streams according to sound features and time frequency. In the schema-based processing, prior knowledge such as language, music, other auditory memory, and endogenous attention helps to compare the auditory input signals with previous experience (Shinn-Cunningham, 2008) (see **Figure 2a**). In laboratory studies, psychologists adopt the dichotic listening paradigm to mimic the Cocktail Party problem. During the task, participants are asked to attend to the auditory materials presented to one ear and ignore the auditory materials presented to the other ear. Afterwards, participants are asked to report the information from the attended or unattended ear. Previous studies show that a higher working memory capacity (WMC) predicts a better attention focus (Colflesh and Conway, 2007; Conway et al., 2001), because a lower capacity cannot accomplish segregation and grouping of any auditory information well (Lotfi et al., 2016). Those findings are in accordance with the controlled attention theory of working memory (Baddeley et al., 1974).

Event-related potential (ERP) N1-P2 components, alpha oscillations, and frequency-following responses (FFRs) disclose how the human brain copes with the Cocktail Party problem (Du et al., 2011; Lewald and Getzmann, 2015; Strauß et al., 2014). The ERP N1 component peaks between 80-120 milliseconds after the onset of a stimulus. It is sensitive to the exogenous auditory stimuli features (Michie et al., 1990). N1 (equivalent in MEG is M100) is generated from the primary auditory cortex (A1) around the superior surface of the temporal lobes (Zouridakis et al., 1998). P2 is always observed as the following component of N1. It peaks at around 200 milliseconds after receiving the external stimulus. These early components support the early selection model of auditory attention (Broadbent, 2013; Lee et al., 2014; Woldorff et al., 1993). Alpha oscillations occur in the parietal cortex and other auditory cortical regions during spatial attention. Selective attention modulates alpha power oscillations in temporal synchrony with the sensory input and enhances the neural activity related to attended stimuli. Wöstmann et al. (2016) conducted a MEG study with a dichotic task and revealed that alpha oscillations are synchronized with speech rates and can predict the listener's speech comprehension. Scalp-recorded frequency-following responses (FFRs) are part of auditory brainstem responses (ABR). They are evoked potentials generated from the brainstem area (Mai et al., 2019). FFRs are phase-locked to the envelope or waveform of the low-frequency periodic auditory stimuli (Zhang and Gong, 2019). In the Cocktail Party problem, FFRs encode important features of speech stimuli to enhance the ability to discriminate the target stimuli from the distracting stimuli (Du et al., 2011). In summary, to exert the auditory selective attention, N1-P2 components are involved in perceiving and detecting the auditory stimuli in the early control processing; alpha oscillations and FFRs are mainly modulated by the selective control to accentuate the target and suppressing noise.

Analogous to the specialized streams of visual selective attention, there are "what" and "where" pathways in the auditory cortex (see **Figure 2b**). The ventral "what" pathway, which involves the anterolateral Heschl' gyrus, anterior superior temporal gyrus, and posterior planum polareactivates, is in charge of identifying auditory objects. The dorsal "where" pathway, which involves the planum temporale and posterior superior temporal gyrus (pSTG), is in charge of spatially localizing auditory objects. Within the "what" pathway, the supratemporal plane-inferior parietal lobule (STP-IPL) network dynamically modulates auditory selective attention; within the "where" pathway, the medial pSTG shows a higher-level representation of auditory localization by integrating the sound-level and timing features of auditory stimuli (Häkkinen and Rinne, 2018; Higgins et al., 2017). In addition, the "where" pathway is observed to activate around 30ms earlier than the "what" pathway implying that top-down spatial information may modulate the auditory object perception (Ahveninen et al., 2006; Alain et al., 2001). However, current studies find that functional overlaps exist in brain areas under different processing pathways, suggesting that brain areas are not function-specific (Schadwinkel and Gutschalk, 2010; Yin et al., 2014). The observed brain activities are not only stimulus-dependent but also task-dependent (Häkkinen, Ovaska, & Rinne, 2015). Besides, a suggested "when" pathway for temporal perception (Lu et al., 2017) deserves to be studied further because the temporal coherence is crucial for binding and segregating features into speech and speaker recognition when attention is engaged. Apart from the paralleled pathways, the distributed processing under different structures may also provide feedback to facilitate the auditory attention (Bizley and Cohen, 2013).

For the Cocktail Party problem, previous neural findings show the attentional selective mechanism occurs in different phases of information processing. Ding et al. (2012) found that the selective mechanism exists in both top-down modulation and bottom-up adaptation during the Cocktail Party problem. When the unattended speech signals were physically stronger, attended speech could still dominate the posterior auditory cortex responses by the top-down execution. Besides, when the intensity of the target was more than 8dB louder than the background, the bottom-up neural responses only adjusted to the target speaker rather than the background speaker. Golumbic et al. (2013) demonstrate that the selective mechanism happens only in the high-level cortices such as the inferior frontal cortex, anterior and inferior temporal cortex, and IPL. Here, only attended speech was selectively retained. However, in the low-level auditory cortices like the STG, both attended and unattended speech were represented. In addition, one study used functional near-infrared spectroscopy (fNIRS)-hyperscanning and found that the brain-to-brain interpersonal neural synchronization (INS) selectively enhances at the left TPJ only between the listener and the attended speaker but not between the listener and the unattended speaker. The listener's brain activity overtakes the speaker's showing a faster speech prediction by the listener. Besides, the INS increased only for the noisy naturalistic conversations with competing speech but not for the two-person conversation and was only associated with the speech content. Their findings implied that the prediction of the speaker's speech content might play an important role in the Cocktail Party Effect (Dai et al., 2018). In sum, the human brain's auditory processing during the Cocktail Party problem is not hierarchical but heterarchical including interactions between different pathways, orders, and adaptations according to the environments (Bizley and Cohen, 2013; Shinn-Cunningham, 2008).

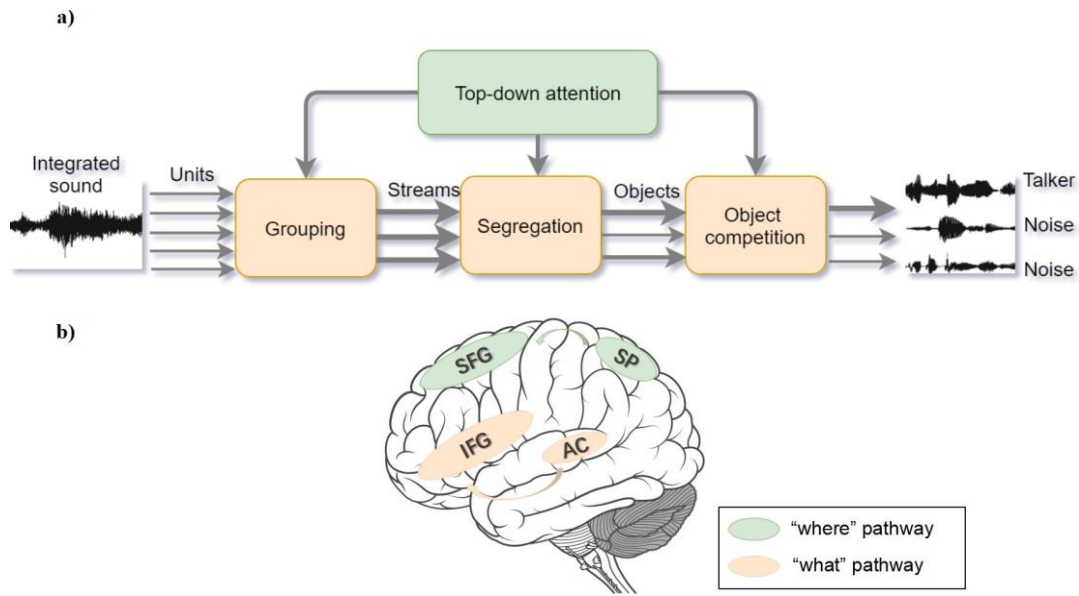

**Figure 2. a)** Auditory selective attention model with interaction between bottom-up processing and top-down modulation. The compound sound enters the bottom-up processing in the form of segregated units and then the units are grouped into streams. After segregation and competition, foreground sound stands out from the background noise. The wider arrow represents the salient object with higher attentional weights. Top-down attention control can modulate processing on each stage (adapted from Shinn (2008) and Bregman (1994)). **b)** The "where" and "what" cortical pathways of auditory attention processing. Within the dorsal "where" pathway, the superior frontal gyrus (SFG) and superior parietal (SP) areas activate during sound localization. Within the ventral "what" pathway, inferior frontal gyrus (IFG) and auditory cortex activate to recognize the object (adapted from Alain et al. (2001)).

## 4.2 Computational models

In the future, we may have moving robots offering food and drinks in noisy restaurants by precisely localizing speaking customers. Steps to solve the Cocktail Party problem in computer science can be mainly separated into: speech separation, sound localization, speaker identification, and speech recognition. The aims of acoustic models for the Cocktail Party problem are: identifying multiple speakers and disentangling each speech stream from noisy background. Numerous classical acoustic models are data-driven and based on algorithms of signal processing (Dávila-Chacón et al., 2018). Those models are robust and with good accuracy but lack the prior knowledge, biological plausibility and rely on the large datasets. Currently, models inspired by the human auditory attention system rely on smaller datasets and have shown improved adaptation. In this section, we focus on the following bio-inspired models: 1) computational auditory scene analysis (CASA): neural oscillator models as examples; 2) saliency models; 3) top-down- and bottom-up-based models.

Based on the Gestalt framework (Rock and Palmer, 1990), the goal of most CASA models is to segregate sounds with similar patterns or connections and group them into independent streams from the mixed auditory scene. Stemming from CASA models, neural oscillator models show good adaptation in auditory segregation. Neural oscillator models perform stream segregation based on the oscillatory correlation. Attention interest is modeled as a Gaussian distribution over then attended frequency. The attentional leaky integrator (ALI) consists of the connection weights between oscillators and the attentional process. The synchronized oscillators activate the ALI to separate sounds into streams like the endogenous attention mechanism (Wrigley and Brown, 2004). Furthermore, to make use of the temporal proximity of sound sources, Wang and Chang (2008) propose a two-dimensional (time and frequency) network oscillator model with relaxation oscillators of local excitation and global inhibition (see Locally Excitatory, Globally Inhibitory Oscillator Network, LEGION (Wang and Terman, 1995)) (see **Figure 3**). Analogous to humans' neural oscillations, the local excitation mechanism makes oscillators synchronize when they are stimulated by the same stimuli and the global inhibition has an effect on the whole network to make oscillators desynchronize by different stimuli (Dipoppa et al., 2016). In their model, sounds with similar patterns (e.g., close frequency, onset or offset time) tend to be grouped into the same stream. One stream corresponds to an assembly of synchronized neural oscillators. The oscillator models mimic the human selective attentional control and show good adaptation to separate the multi-tone streaming.

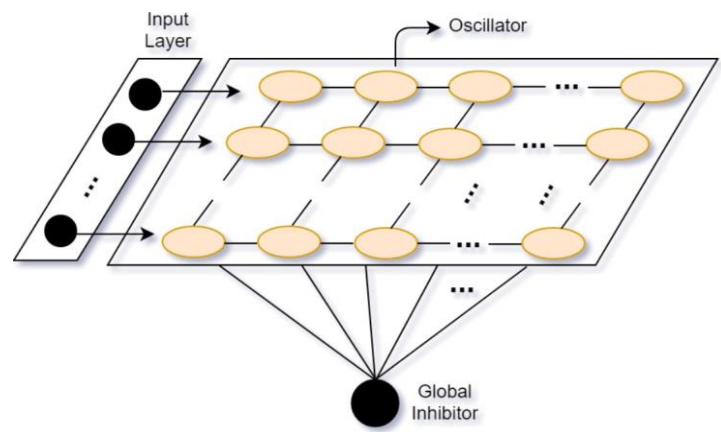

**Figure 3.** Locally Excitatory, Globally Inhibitory Oscillator Network (LEGION) (adapted from Wang and Terman (1995)).

The oscillator models try to mimic the endogenous attentional control while the saliency models try to mimic the exogenous attention. Similar to visual saliency models (see section 3.2), auditory saliency models are built by abstracting features (intensity, frequency contrast, and temporal contrast) from the sound "intensity image", which is a visual conversion of auditory time-frequency spectrograms and normalized to be an integrated saliency map (Kalinli and Narayanan, 2007; Kayser et al., 2005) (see **Figure 4b**). Considering that humans and other primate animals can process the pure auditory signals without any visual conversion, Kaya and Elhilali (2017) modify the auditory saliency model by directly extracting the multi-dimensional temporal auditory signal features (envelope, frequency, rate, bandwidth, and pitch) of the auditory scene as input. Their model relies on the selection of parameters to reduce error rates of the saliency determination by fewer features. To integrate both endogenous and exogenous attention in the model, Morissette and Chartier (2015) propose a model by extracting frequency, amplitude, and position as features and connecting them with the oscillator model LEGION. Segments with consistent features are bound into the saliency map according to the temporal correlation. Notably, a module of inhibition-of-return (IOR) is inserted to inhibit attention from fixing at the most salient scene for a long time. This mechanism can achieve the attentional shifting and orientation (Posner and Cohen, 1984). Several limits exist for developing the auditory saliency models. Firstly, there is no apparent physical marker for a person to locate sounds compared with eye gaze used in visual saliency models. Secondly, metrics to define auditory saliency are not unified (Kaya and Elhilali, 2017). Thirdly, unlike visual attention, acoustic signals distribute across different frequency bands and time windows. This makes auditory models reckon much on the temporal features. Therefore, more high-level features and top-down modulation should be taken into consideration for the model to indicate the significant sound stream.

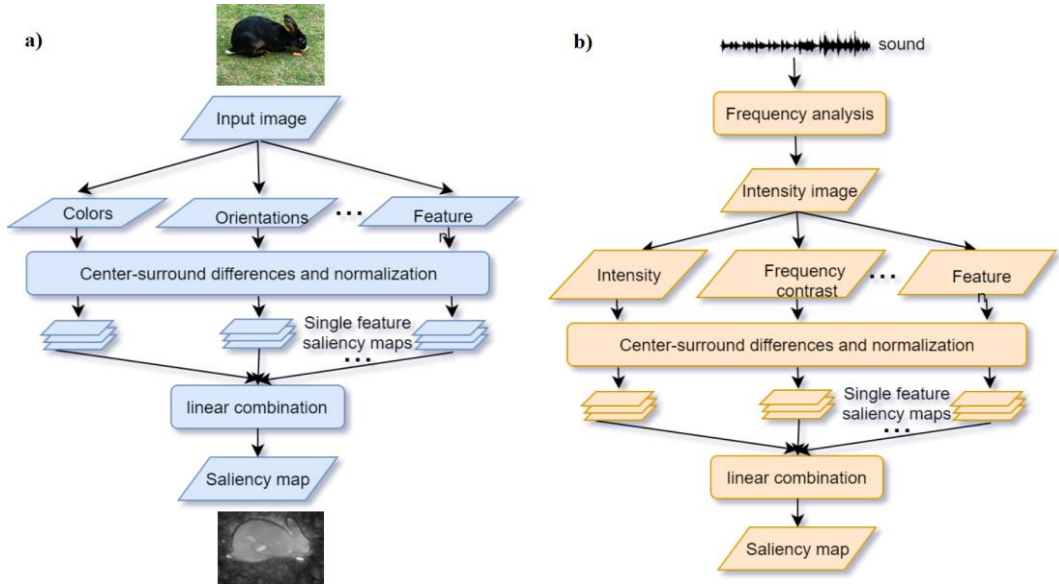

**Figure 4. a)** Visual saliency model. Features are extracted from the input image. The center-surround mechanism and normalization are used to generate the individual feature saliency maps. Finally, the saliency map is generated by a linear combination of different individual saliency maps (adapted from Itti et al. (1998)); **b)** Auditory saliency model. The structure of the model is similar to the visual saliency model by converting sound inputs into a frequency "intensity image" (adapted from Kayser et al. (2005)).

Prior knowledge (e.g., memory, prediction, and expectation) also plays a crucial role in human auditory perception, therefore several top-down- and bottom-up-based models integrate the prior knowledge into the data-driven models. Some of them extract acoustic features of the target sound and store them in memory-like modules to mimic humans' long-term memory function as top-down modulation. Oldoni et al. (2013) combine a self-organized map (SOM) of the acoustic features in the bottom-up processing to continuously learn the saliency and novelty of acoustic features. After training, each SOM unit matches up with a representative sound prototype. For the top-down processing, the IOR and LEGION mechanisms are introduced to shift and select attention, respectively. Xu et al. (2015) propose an Auditory Selection framework with Attention and Memory (ASAM). In this model, there is one speech perceptor extracting the voiceprint of speakers and accumulating the voiceprint in a lifelong-learning memory module during the training phase to be the prior knowledge for the model. Later, the learned voiceprint is used to attend and filter the target speech from the sound input to achieve the top-down and bottom-up interaction. The testing performance showed good robustness and adaptation for both top-down (follow a specific conversation) and bottom-up (capture the salient sound or speech) attention tasks.

Shi et al. (2018) propose the Top-Down Auditory model (TDAA) and use the prediction of the target speaker as the top-down modulation. Their model contributes to the auditory scene analysis with multiple unknown speakers. They adopt the Short-Time Fourier Transformation (STFT) and Bidirectional Long-Short Term Memory (BiLSTM) to predict the number of the speakers. Later, the classifier recurrent neural networks (RNN) separate the most salient speaker and iterate until no more speakers can be separated to avoid repeated prediction. Finally, an attention module is used to separate each speaker's spectrum from the spectrum mixture. Besides, binaural models are apt to make use of the spatial localization information to address the Cocktail Party problem. For instance, Ma et al. (2018) train DNNs to localize acoustic features in full 360° azimuth angles. After the training phase, the binaural localization with spectral features is used as prior knowledge in the top-down modulation of the model. Their model serves to predict the speech with different localizations under noisy situations with room reverberation. In summary, those top-down and bottom-up interaction models incorporate mechanisms of processing in the human auditory system. They selectively attend or shift attention to the target speech dynamically rather than only focusing on the stream separation, which can be more adaptive to those uncertain and complex auditory scenarios.

## 5 AUDIOVISUAL CROSSMODAL SELECTIVE ATTENTION

### 5.1 Behavioral and neural mechanisms

In order to survive in an uncertain and multimodal world, humans develop the ability to integrate and discriminate simultaneous signals from multiple sensory modalities, such as vision, audition, tactile, and olfaction. For example, humans can make use of visual cues like lip movement and body gestures to recognize and localize sounds in noisy circumstances. The crossmodal integration ability is beneficial for humans to localize and perceive objects but can also cause ambiguity. Crossmodal conflicts arise when information from different modalities are incompatible with each other and can result in failures of the crossmodal integration and object recognition. To resolve conflicts, selective attention is required to focus on the task-relevant modality information and to ignore the interference from irrelevant modalities (Veen and Carter, 2006). For instance, due to the underdevelopment of executive function and conflict control, Autism Spectrum Condition (ASC) participants showed poorer performance on suppressing irrelevant modalities than normal participants during crossmodal tasks (Chan et al., 2016; Poole et al., 2018). In this subsection, we mainly talk about behavioral and neural mechanisms of selective attention underlying audiovisual crossmodal integration and conflict resolution.

First, how and when does a crossmodal conflict occur? Previous studies proved that humans tend to integrate visual and auditory stimuli with spatial-temporal linkage into the same object (Senkowski et al., 2008). The "Unity assumption" proposes that when humans believe that the multisensory inputs they perceive are generated from the same source, crossmodal integration occurs (e.g. when students think the speech they hear in the lecture room matches the lip movements of the professor, they believe that the speech is from the professor) (Roseboom et al., 2013). Besides, prior knowledge and experience can generate expectation effects to facilitate object recognition during crossmodal integration. Therefore, when the stimuli from different modalities are spatially (e.g., ventriloquism effect (Choe et al., 1975)) or temporally incongruent (e.g., double flash illusion (Roseboom et al., 2013)) or contrary to our expectations (e.g., see a cat with a "bark" sound), humans perceive crossmodal conflicts. During the early integration stage, selective attention plays the role of capturing the salient visual and auditory stimuli by bottom-up processing. When conflicts are detected, selective attention executes a top-down modulation from higher-level semantic representations according to the internal goal and relevant modalities. The crossmodal information processing is not only a feed-forward process but also contains backward feedback and recurrent processes, which are important to facilitate the primary sensory processing (Talsma et al., 2010) (see **Figure 5a**).

Second, which modality dominates when humans are confronted with audiovisual conflicts? Lots of studies have examined the "ventriloquism effect", which originally refers to the strong visual bias during the sound localization (Choe et al., 1975; Thurlow and Jack, 1973; Warren et al., 1981). Research findings show that this strong modality bias changes through the lifespan of a human (Sloutsky, 2003). Compared to toddlers, adults are more likely to have visual stimuli preferences (Sloutsky, 2003). Some researchers argue that the ventriloquism effect results from an optimal or suboptimal decision-making strategy, especially when unimodal stimuli are blurred. If the auditory stimuli are more reliable than the visual stimuli, an auditory bias occurs as well (Roseboom et al., 2013; Alais and Burr, 2004; Ma, 2012; Shams and Kim, 2010). To sum up, vision in general has a higher spatial resolution than audition, whereas audition has a higher temporal resolution than vision. As the modality appropriateness hypothesis points out, the information from one modality dominates according to the temporal or spatial features of the audiovisual event and the modality with the higher accuracy (Welch and Warren, 1980).

Third, how do humans resolve crossmodal conflicts? In the conflict-monitory theory, the module of conflict monitoring (CM) is activated when conflicts are detected and passes the signal to the executive control (EC) module to accomplish the task-related conflict resolution by the top-down attentional control (Botvinick et al., 2001). From the previous findings, to perceive crossmodal signals and detect crossmodal conflicts, selective attention plays the role of gating crossmodal coupling between sensory function areas in a modality-general fashion (Convento et al., 2018; Eimer and Driver, 2001; Mcdonald et al., 2003). However, to solve crossmodal conflicts, selective attention inclines towards processing in a modality-specific fashion (Mengotti et al., 2018; Yang et al., 2017).

Except for some specific vision and audition processing brain areas, the superior colliculus (SC) is a crucial brain area with multisensory convergence zones from visual and auditory primary cortices to higher-level multisensory areas. As it is mentioned in section 3.1, the SC also implements selective attention by orienting both covert and overt attention towards the salient stimulus and triggers corresponding motor outputs (e.g., eye movements, saccades) (Krauzlis et al., 2013; Meredith, 2002; Wallace et al., 1998). Besides, the superior temporal sulcus (STS), inferior parietal sulcus (IPS), frontal cortex (including premotor and ACC), and posterior insula are involved in the crossmodal processing (for review see (Calvert, 2001; Stein and Stanford, 2008)). Within the crossmodal brain functional network, the STS plays the role of linking unimodal representations (Hertz and Amedi, 2014). The parietal lobe is thought to process representations of visual, auditory, and crossmodal spatial attention (Farah et al., 1989). However, when audiovisual inputs are incongruent, crossmodal attenuations or deactivations occur (Kuchinsky et al., 2012). To resolve conflicts, as human fMRI studies have shown, the dorsal anterior cingulate cortex (dACC) is responsible for dealing with conflicts between the current goal and irrelevant distractors. The dACC is positively correlated with attention orientation and interference suppression (Weissman et al., 2004). Song et al. (2017) conducted a mice experiment by using a task with audiovisual conflicts, where audition was required to dominate vision. They found that when the conflict occurred, the co-activation of the primary visual and auditory cortices suppressed the response evoked by vision but maintained the response evoked by audition in the posterior parietal cortex (PPC).

Electrophysiological studies have shown the existence of cells that respond to stimulation in more than one modality to accomplish crossmodal integration and conflict resolution. Diehl et al. (2014) found that neurons in the ventrolateral prefrontal cortex (VLPFC) of Macaques were bimodal and nonlinear multisensory. When incongruent faces and vocalizations were presented, those neurons showed significant changes with an early suppression and a late enhancement during the stimulus displaying period. Other experimental evidence argues that coherent oscillations across different modality cortices are the key mechanism of the crossmodal interplay (Wang, 2010). An enhancement of the phase locking for the short-latency gamma-band activity (GBA) is found for the attended multisensory stimuli. The early GBA enhancement enables the amplification and integration of crossmodal task-relevant inputs (Senkowski et al., 2008). Incongruent crossmodal inputs cause a stronger gamma-band coherence than congruent inputs suggesting the involvement of gamma oscillations decoupling under crossmodal binding (Misselhorn et al., 2019). Attentional control during the crossmodal integration and conflict resolution is associated with alpha-band effects from the frontoparietal attention network rather than primary sensory cortices. Frontal alpha oscillations are involved in the top-down perceptual regulation; parietal oscillations are involved in the intersensory reorientation (Misselhorn et al., 2019). Reversed to the gamma oscillation patterns, incongruent conditions showed weaker alpha oscillation changes compared to congruent conditions. This gamma-alpha oscillation cycle pattern is proposed to be the information gating mechanism by inhibiting task-irrelevant regions and selectively routing the task-relevant regions (Bonnefond and Jensen, 2015; Jensen and Mazaheri, 2010). In sum, the convergent brain areas accomplish the crossmodal integration and conflict resolution by processing multimodal projections from visual and auditory primary cortices. Neural oscillations coordinate the temporal synchronization between the visual and auditory modality.

## 5.2 Computational models

Compared with unimodal learning, crossmodal learning is more beneficial to model complex behaviors or achieve high-level functions on artificial systems, such as object detection (Li et al., 2019), scene understanding (Aytar et al., 2017), lip reading (Mroueh et al., 2015; Chung et al., 2017), etc. In psychology, crossmodal learning (cml) research focuses on how crossmodal learning helps humans to recognize objects or events by integrating multimodal information and eliminating the crossmodal ambiguity (Calvert, 2001). In computer science, cml research focused on recognizing one modality by using a multimodal dataset or making use of the data from one single modality and retrieve relevant data of other modalities (Wang et al., 2017). Nowadays, cml in computer science addresses any kind of learning that involves information obtained from more than one sensory or stimulus modality (Skocaj et al., 2012). In robotics, cml research focuses on multisensory binding and conflict resolution to make robots interact with the environment with higher robustness and accuracy. In this section, we mainly introduce the computational modeling work on selective attention from the audiovisual crossmodal perspective.

Many studies focus on multimodal fusion (Ramachandram and Taylor, 2017), but research about selective attention and conflict resolution in computer science is limited. Parisi et al. conducted a series of audiovisual crossmodal conflict experiments to explore conflict resolution patterns of humans (Parisi et al., 2017, 2018; Fu et al., 2018). During human behavioral tasks, visual and auditory stimuli were presented in an immersive environment. Four loudspeakers were set behind the corresponding positions on a 180-degree screen, where four human-like avatars with visual cues (lip movement or arm movement) were shown. The visual cue and the sound localization could be congruent or incongruent (e.g., the left-most sound with the right-most avatar's lip movement). During each trial, human participants were asked to determine where the sound was coming from. Analyses of human behavior results showed that the congruence effect exists stably under all the incongruent audiovisual conditions. Even though arm moving was visually more salient than lip moving, humans had higher error rates of the sound localization when viewing lip movement. This suggests that lip moving might contain more speech or semantic information so it is more difficult to be ignored. Besides, the magnitude of the visual bias was also significant when the incongruent AV stimuli were coming from the two avatars at the extreme right and left sides of the screen. This indicated a wider integration window than other simplified scenes. Based on the bio-inspired cortico-collicular architecture, deep and self-organizing neural networks consisting of visual and auditory neuron layers and crossmodal neuron layers were used to learn crossmodal integration and conflict resolution (see **Figure 5b**). In this way, human-like responses were modelled and embedded in an iCub robot.

Except for the perceptual level of modeling, another approach encoded the expectation learning effect and unity assumption into DNNs to learn semantic-level crossmodal representations (e.g., a dog image associated with a bark sound as congruent and with a meow sound as incongruent) (Barros et al., 2018). Within the model architecture, separate perception modalities learnt the novel concepts (bottom-up) and a self-organizing layer learnt the concurrent modality representations (top-down). The work above shows that DL can simulate humans' selective attention and conflict resolution on audiovisual sound localization and semantic association. Due to the limited resources and sensory modules, the future exploration of modeling and simulating conflict processing is desirable in crossmodal robotics. Besides, conflict resolution mechanisms can boost the applicability and accuracy on robots in real human-robot interaction scenarios. Robots can select the more reliable modalities and reduce the distraction and errors. For humans, the capacity of conflict adaptation plays a crucial role in learning and adapting to the environment. When human toddlers detect any conflict between the current environment and their prior knowledge, they will generate curiosity and be motivated to learn new knowledge or rules. Curiosity is also important for the trial and error learning of robots (Hafez et al., 2019). Therefore, selective attention and conflict processing modules need to be integrated in intelligent systems to prioritize the response and promote the learning progress.

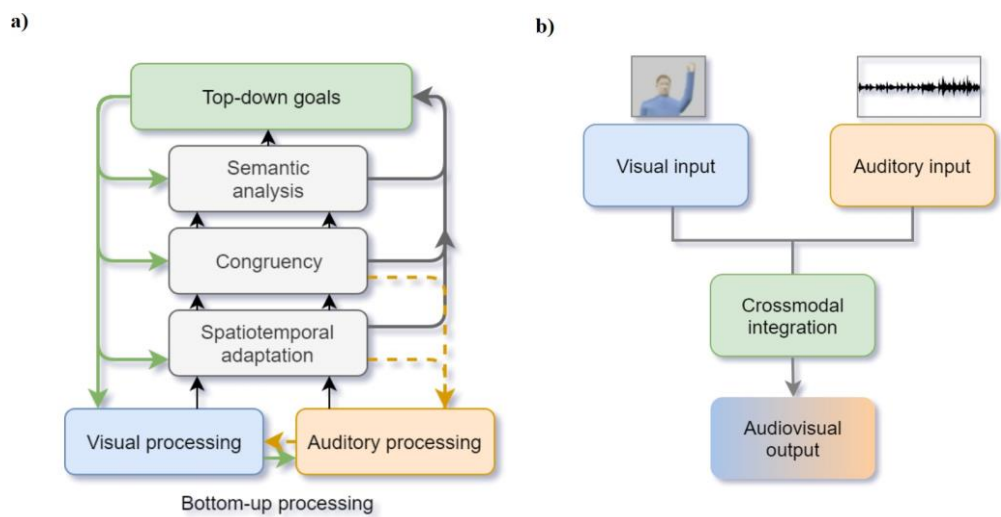

**Figure 5. a)** Human crossmodal integration and attentional control. The black and grey arrows denote the feed-forward bottom-up stimulus saliency processing and the green arrows denote the top-down modulation of attention. The yellow dashed arrows represent the recurrent adjustment (adapted from Talsma et al. (2010)); **b)** Artificial neural networks of crossmodal integration. The crossmodal integration mechanisms are used to realign the input from visual and auditory modalities (adapted from Parisi et al. (2017; 2018)).

# 6 CONCLUDING REMARKS AND OUTSTANDING QUESTIONS

The current review summarizes experimental findings, theories, and model approaches of audiovisual unimodal and crossmodal selective attention from psychology, neuroscience, and computer science perspective. Currently, psychologists and neural scientists are working towards computational modeling, standardizing, and replication. In parallel, computer scientists are trying to design and make agent systems more intelligent with higher-level cognitive functions, meta-learning abilities, and lower learning costs. Some advantages, unresolved problems, and future directions of collaborative research in psychology, neuroscience, and computer science are summarized as follows:

Advantages. One the one hand, findings and methods from psychology and neuroscience can interpret and improve models' performance (Hohman et al., 2018). For instance, representational similarity analysis (RSA) is nowadays also used to compare the responses recorded in fMRIs and artificial systems like deep learning CNNs. RSA analyzes the similarity of fMRI responses and brain representations by a set of stimuli (Kriegeskorte et al., 2008). Dwivedi et al. (2019) found that RSA shows good performance on transfer learning and task taxonomy by computing correlations between the models on certain tasks. On the other hand, the-state-of-the-art approaches offer tools to analyze big data of neural findings. For example, the SyConn framework used deep CNNs and random forest classifiers to accelerate data analyses on animal brains to compute the synaptic wiring of brain areas (Dorkenwald et al., 2017). Another potential application of computational modeling is examining theories and interpreting mechanisms in neuroscience. Models can be built to simulate normal behaviors and then mimic the "damage" by removing units of the models. If the "damage" causes similar abnormal behaviors like psychiatric patients do, the removed units may be the corresponding mechanisms to the behaviors.

Unresolved problems. Even though we have reviewed and summarized a number of findings from psychology and computer science, lots of unsolved issues of attention processing remain to be disclosed. The simulation work of cognitive control and conflict processing is insufficient on robots. Besides, the problem of perceptual consistency has not been deeply addressed in computer science. For humans, it is easy to recognize one object from different perspectives, such as finding an open door in a dim room. Moreover, humans can transfer the intrinsic knowledge to learn and infer new objects or concepts with a small number of learning samples. However, artificial intelligent systems cannot reach humans' performance yet. For example, even though the scale-invariant feature transform (SIFT) algorithm (Lowe et al., 1999) can extract features from variant shapes of the same object, it cannot recognize the variant objects when only colors exist without any structural patterns. Current deep learning approaches like the VGG net (Simonyan and Zisserman, 2015) has shown better performance on object recognition than traditional approaches. However, such deep networks rely on the training dataset and need substantial computational resources.

Future directions. There is a lot of potential for psychologists and computer scientists to work together to investigate both human cognition and intelligent systems. On the one hand, psychologists can focus on designing paradigms to diagnose and remedy shortages of current models to improve the model accuracy. Besides, neural studies are still needed to understand human brain mechanisms better. It will be insightful to develop bio-inspired computational models with a better interpretability. On the other hand, for computer science, enhancing the complexity of models to increase the adaptivity and flexibility to the environment is required. At last, to balance the computational complexity and biological plausibility is also crucial, because humans' behavioral patterns are limited by their capacity and energy load, even though properties of machines will keep improving. In summary, deepening the understanding of each processing mechanism rather than only describing phenomena is the direction for research from both sides to endeavor.

## CONFLICT OF INTEREST STATEMENT

The authors declare that the research was conducted in the absence of any commercial or financial relationships that could be construed as a potential conflict of interest.

*Fu et al.*                Selective attention mechanisms and modeling*Fu et al.* Selective attention mechanisms and modeling
## AUTHOR CONTRIBUTIONS

SW and XL contributed to the conception and organization of the manuscript. DF wrote the first draft of the manuscript. All authors contributed to manuscript reading and revision and approved the submitted version.

## FUNDING

This work is supported by National Natural Science Foundation of China (NSFC), the German Research Foundation (DFG) under project Transregio Crossmodal Learning (TRR 169), and CAS-DAAD joint fellowship.

## ACKNOWLEDGMENTS

We especially appreciate feedback and support from the Knowledge Technology Group in Hamburg and the Carelab in Beijing. We thank Katja Kösters for proofreading the manuscript and improving the language clarity. We also thank Dr. German I. Parisi, Ge Gao, Zhenghan Li, Honghui Xu, and Antonio Andriella for the fruitful discussions.